\pdfoutput=1

\documentclass[11pt]{article}

\usepackage[]{acl2022}

\usepackage{times}
\usepackage{latexsym}

\usepackage[T1]{fontenc}

\usepackage[utf8]{inputenc}

\usepackage{microtype}
\usepackage{algorithm}
\usepackage{multicol}
\usepackage{multirow}
\usepackage{algorithmicx}
\usepackage{mathrsfs}
\usepackage{graphicx}
\usepackage{float}
\usepackage{amssymb}
\usepackage{amsmath, bm}
\usepackage{array}
\usepackage{algpseudocode}

\usepackage{xcolor}

\title{Type-enriched Hierarchical Contrastive Strategy for \\ Fine-Grained Entity Typing}

\author{Xinyu Zuo, Haijin Liang, Ning Jing, Shuang Zeng, Zhou Fang and Yu Luo
 \\
	 Tencent Inc.  \\
	{\tt \{xylonzuo,hodgeliang,shuangzeng,akirafang,yamiluo\}@tencent.com} \\
	{\tt ning.jing.ustc@gmail.com}
	} 

\begin{document}
\maketitle
\begin{abstract}
Fine-grained entity typing  (FET) aims to deduce specific semantic types of the entity mentions in text. Modern methods for FET mainly focus on learning what a certain type looks like. And few works directly model the type differences, that is, let models know the extent that one type is different from others. To alleviate this problem, we propose a type-enriched hierarchical contrastive strategy for FET. Our method can directly model the differences between hierarchical types and improve the ability to distinguish multi-grained similar types. On the one hand, we embed type into entity contexts to make type information directly perceptible. On the other hand, we design a constrained contrastive strategy on the hierarchical structure to directly model the type differences, which can simultaneously perceive the distinguishability between types at different granularity. Experimental results on three benchmarks, BBN, OntoNotes, and FIGER show that our method achieves significant performance on FET by effectively modeling type differences.

\end{abstract}

\section{Introduction}
\label{sec:intro}
\begin{figure}[t]\
	\centering
	\includegraphics*[clip=true,width=0.48\textwidth,height=0.25\textheight]{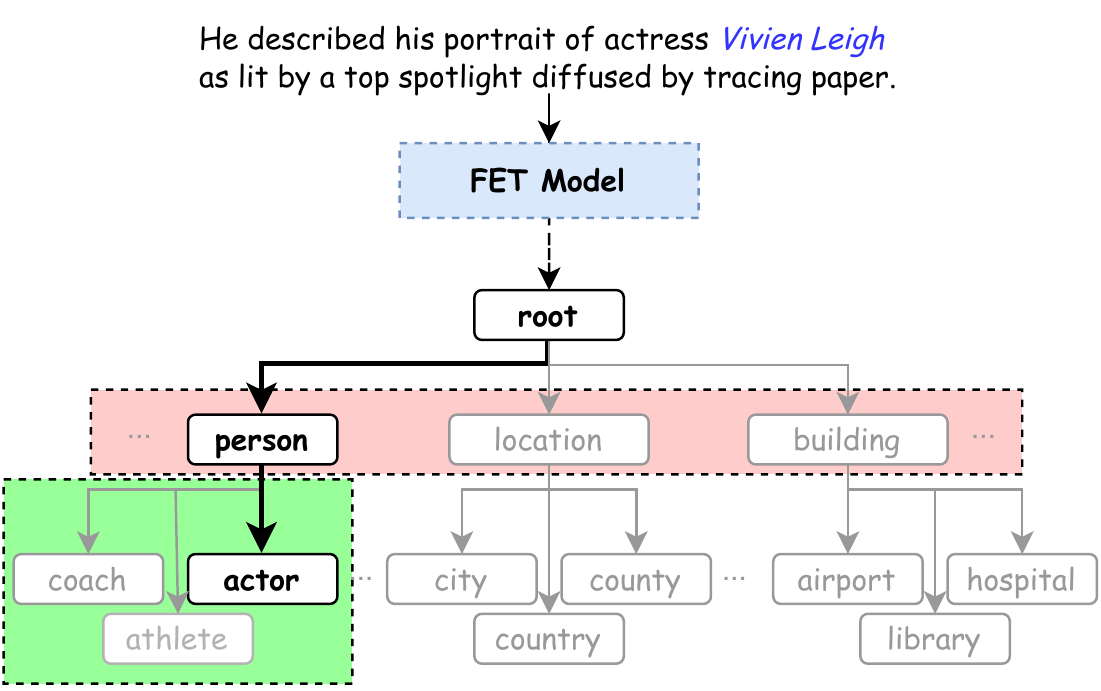}
	\caption{Example of fine-grained entity typing based on FIGER ontology. Green Box: the scope of visible types about \emph{"person"} of the fine-grained contrastive strategy. Red Box: the scope of visible types of the coarse-grained contrastive strategy.} \label{fig1}
\end{figure}

Entity typing is a fundamental research problem in natural language processing (NLP), which aims to deduce the semantic types of the entity mentions in text. With the deepening of text understanding, the type sets of entities become more refined and ranging in from dozens \cite{hovy2006ontonotes} to hundreds \cite{weischedel2005bbn,ling2012fine} or thousands \cite{choi-etal-2018-ultra}. Therefore, fine-grained entity typing (FET) has gained more attention, which focuses on assigning more specific types to entities. For sentence in Figure \ref{fig1}, a FET system needs to assign a coarse-grained type \emph{"/person"} and a fine-grained type \emph{"/person/actor"} to the entity \emph{"Vivien Leigh"}. The inferred fine-grained types could provide more specific prior knowledge for downstream NLP tasks, such as question answering \cite{lee2006fine} and entity linking \cite{leszczynski2022tabi}.

Considering the partial ontology of the FIGER dataset \cite{ling2012fine} in Figure \ref{fig1}, fine-grained entity types are often linked together in a hierarchical taxonomy, which makes the type boundaries increasingly blurred, especially for sub-types under the same coarse type. As shown in Figure \ref{fig1}, the fine-grained types \emph{"coach"}, \emph{"athlete"} and \emph{"actor"}, all of which fall into the coarse-grained type \emph{"person"}, are less differentiated.

In order to identify fine-grained types, prior work has concentrated on excavating more informative representations of types or entities, which benefiting from hand-crafted features \cite{ren-etal-2016-afet}, external resources \cite{onoe2020fine,li2022ultra} or external pre-trained task \cite{xu2020syntax}. Most of them focus on learning \emph{what a certain type looks like}, but few works have gone further to directly model the differences between types, that is, let models know \emph{the extent that one type is different from others}, which is more effective to distinguish among similar fine-grained types.

How to directly model the type differences? We argue that there are two key points, \emph{Entity Type Awareness}, which refers to directly perceiving what type of entity is in the sentence, and \emph{Type Differences Measure}, which refers to modeling how different the perceived type is from other types. For the first point, the intuitive idea is to expose the types directly, leading to a direct focus on what type the context represents. For the second point, heuristically, \emph{direct is effective}, i.e., directly modeling which contexts represent the same types and which are different is the most efficient way to measure differences. 

To this end, we propose a ty\textbf{P}e-enriched h\textbf{I}erarchical \textbf{CO}ntrastive stra\textbf{T}ey (\textbf{PICOT}) for fine-grained entity typing. Specifically, for \emph{entity type awareness} with limited annotated data, inspired by prompt learning in entity typing \cite{ding2021prompt}, PICOT embeds the entity types in contexts via prompts to build type-rich expressions that guide the learning of correct types. Additionally, for \emph{type differences measure}, PICOT takes a constrained contrastive strategy on hierarchical taxonomy to directly model the type differences from type-rich expressions. Concretely, as shown in Figure \ref{fig1}, PICOT is only concerned with the fine-grained types under the same coarse-grained type to learn the differences between fine-grained types. Similarly, PICOT is not concerned with what the fine-grained types are when distinguishing dissimilarities between coarse-grained types. Methodologically, PICOT learns the type differences at different granularity through type-rich expressions by limiting the scope of attention to types. Moreover, to further show models what a particular type is, we introduce a small number of type descriptions that directly expose richer type knowledge.

In experiments, we evaluate our model on three benchmarks. First, we concern with the standard evaluations and show that our model achieves the state-of-the-art performance on FET. Then we estimate the main components of PICOT. Finally, we do a visual analysis of the effectiveness of the type differentiation of PICOT.

In summary, the contributions are as follows:
\begin{itemize}
	\item We propose a type-enriched hierarchical contrastive strategy (\textbf{PICOT}) for fine-grained entity typing. Our method can directly model the differences between hierarchical types and improve the ability to distinguish multi-grained similar types.
		
	\item First, we embed types into entity contexts to make type information directly perceptible. Then we design a constrained contrastive strategy on hierarchical taxonomy to directly model type differences at different granularities simultaneously.
	
	\item Experimental results on three benchmarks show that PICOT can achieve the SOTA performance on FET with limited annotated data.
\end{itemize}

\section{Related Work}
\label{sec:related}
\paragraph{Entity Typing}
Named entity recognition \cite{tjong-kim-sang-de-meulder-2003-introduction} and entity typing \cite{ling2012fine,gillick2014context} are fundamental research problems in NLP. Recently researchers pay more attention to fine-grained entity typing (FET) and ultra-fine entity typing (UFET) \cite{choi-etal-2018-ultra}, which predict specific fine or ultra-fine types for given entities. To do so, obtaining more labeled data is the first research perspective, represented by distant supervision \cite{ling2012fine,chen2019improving}. With these, some researchers had focused on how to reduce noises in automatically labeled data \cite{gillick2014context,ren-etal-2016-afet,ren2020fine,wu2019modeling,pan2022automatic,zhang2021learning,pang-etal-2022-divide}. Additionally, another key challenge is how to deal with hierarchical ontology. Most prior works regarded the hierarchical typing problem as a multi-label classification task and incorporated the hierarchical structure in different ways \cite{ren-etal-2016-afet,shimaoka-etal-2017-neural,xu2018neural,murty-etal-2018-hierarchical,chen-etal-2020-hierarchical,chen-etal-2022-learning-sibling}. 
 
Some works attempted to mine more label information or better label representation. \citet{abhishek-etal-2017-fine} enhanced the label representation by sharing parameters; \citet{lopez-strube-2020-fully} embed types into a high-dimension; \citet{xiong-etal-2019-imposing} introduced associated labels to enhance the label representation;  \citet{rabinovich-klein-2017-fine,lin-ji-2019-attentive} exploited co-occurrence structures and latent label representation; Additionally, several novel textual representations were applied to obtain richer entity contextual information, such as prompt based architecture \cite{ding2021prompt} and box embeddings framework \cite{onoe-etal-2021-modeling}.

Moreover, FET and UFET suffer from an obvious issue of the unseen types due to the lack of annotated data. Therefore, a variety of paradigms were be studied to alleviate this issue \cite{huang2016building,ma-etal-2016-label,obeidat-etal-2019-description,zhou-etal-2018-zero,zhang-etal-2020-mzet,chen-etal-2021-empirical}. Moreover, some works further drew on different large-scale external data or knowledge to understand entity types \cite{onoe2020fine,xu-etal-2021-syntax,dai-etal-2021-ultra,li2022ultra}. 

In summary, few prior works focus on directly modeling type differences. Therefore, this paper tries to let models know that one type is different from others without large-scale external resources. See Appendix B for more details.

\paragraph{Contrastive Learning}
Contrastive learning aims to further improve the model's ability to distinguish positive and negative examples, and has been a popular method for representation learning on computer vision tasks \cite{hjelm2018learning,pmlr-v119-chen20j,he2020momentum}. Recently, some researches have applied contrastive learning to natural language understanding tasks, aiming to obtain better text representations or to distinguish similar labels, such as the event causality identifier \citep{zuo-etal-2021-improving}, the contrastive self-supervised encoder \citep{DBLP:journals/corr/abs-2005-12766}, the supporting clustering framework \cite{DBLP:journals/corr/abs-2103-12953}, the abstractive summarization framework \cite{liu-liu-2021-simcls}, the contrastive fine-tuning paradigm of pre-trained language for fine-grained text classification \cite{suresh-ong-2021-negatives}, and so on. In this paper, we propose a constrained contrastive framework to directly model the hierarchical type differences.

\paragraph{Prompt Learning}
Prompt learning aims to leverage language prompts as contexts, and downstream tasks can be expressed as some cloze-style objectives similar to those pre-training objectives. Recently, a series of hand-crafted prompts have been widely used in natural language understanding \cite{DBLP:journals/corr/abs-2103-10385,schick-schutze-2021-exploiting,DBLP:journals/corr/abs-1909-00505,DBLP:journals/corr/abs-1909-01066,DBLP:journals/corr/abs-1806-02847,ding2021prompt}. Moreover, to avoid expensive prompt design, automatic prompt has also been explored \cite{ren-etal-2016-afet,DBLP:journals/corr/abs-2010-15980,schick-schutze-2021-exploiting}, and some continuous prompts have also been proposed \cite{DBLP:journals/corr/abs-2104-08691,DBLP:journals/corr/abs-2101-00190}. In this paper, we embed prompts to directly expose types in the entity contexts.

\section{Problem Formulation}
\label{sec:Prob}
The input of fine-grained entity typing (FET) is a dataset $\mathcal{D}=\{x_1, x_2, ..., x_n\}$ with $n$ sentences, a pre-defined hierarchical type ontology $\mathcal{Y}$, and each sentence $x$ contains a marked entity $e$. A FET system is required to assign corresponding types to the given marked entity. Methodologically, for each input sequence $w_n=\{w^1_n, w^2_n, ..., w^t_n\}$ of the sentence $x_n$, FET aims to predict the correct multi-grained types $Y_n=\{y^1_n, y^2_n, ..., y^m_n\} \in \mathcal{Y}$ of the marked entity $e_n=\{w^l_n, ..., w^r_n\}$. For example in Figure \ref{fig1}, the correct type set of \emph{"Vivien Leigh"} is $\{/person, /person/actor\}$, which contains a coarse-grained type "$/person$" and a fine-grained type "$/person/actor$".

\section{Methodology}
\label{sec:method}
\begin{figure}[t]\
	\centering
	\includegraphics*[clip=true,width=0.49\textwidth,height=0.28\textheight]{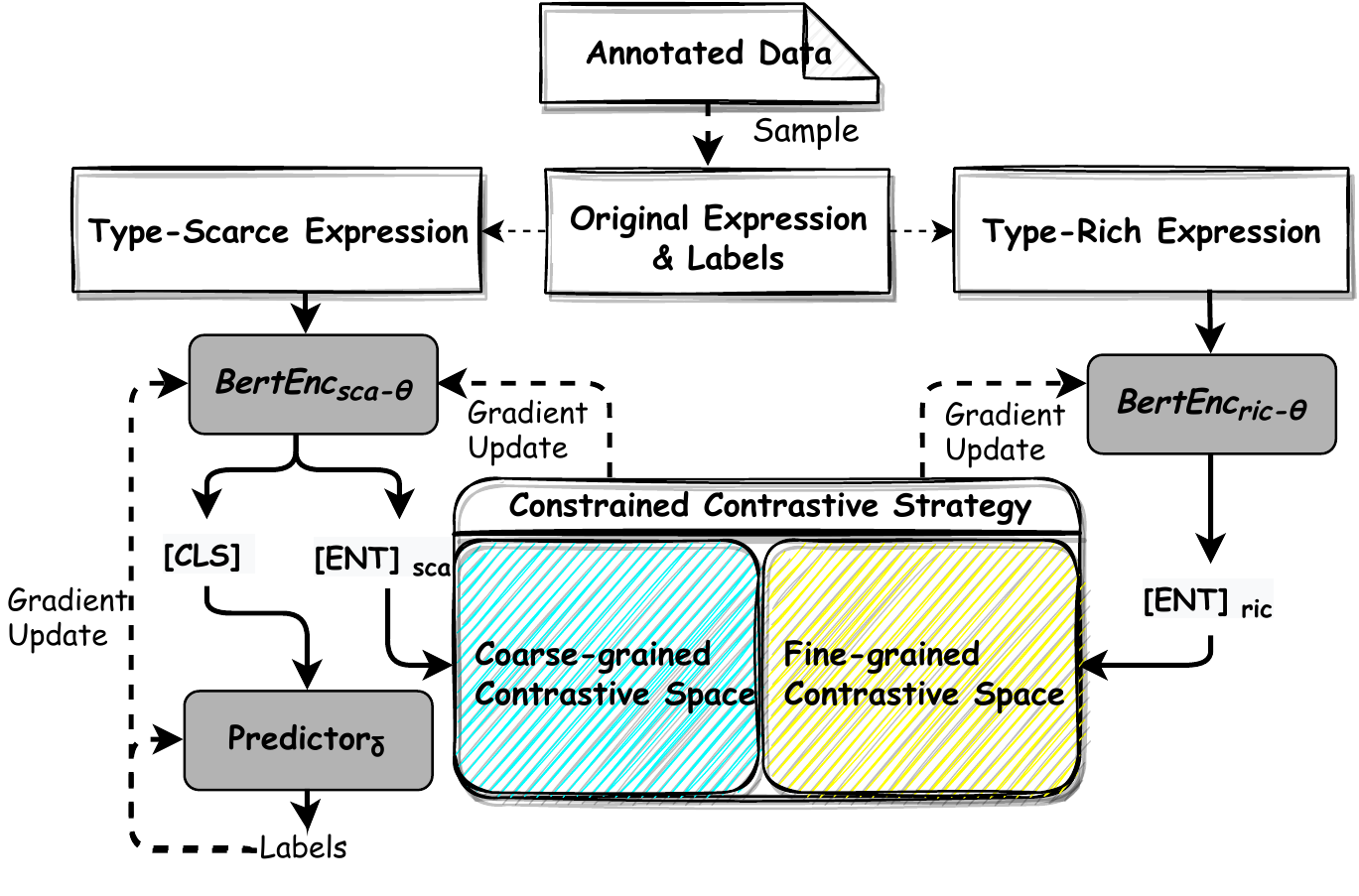}
	\caption{The framework of PICOT for FET (Sec. \ref{sec:method}).} \label{fig2}
\end{figure}

As shown in Figure \ref{fig2}, there are two key stages of PICOT for fine-grained entity typing.

\begin{itemize}
    \item \textbf{Prompt-guided expression construction}  (\textbf{ProExp}, Sec. \ref{sec:ProExp}). For \emph{entity type awareness}, we construct two kinds of prompt-guided expressions, the \emph{type-scarce expression} and the \emph{type-rich expression}, which perceive the type patterns in context and expose type information directly, respectively.
    
    \item \textbf{Contrastive type knowledge transfer} (\textbf{ConTKT}, Sec. \ref{sec:ConTKT}). For \emph{type differences measure}, we propose a \emph{constrained contrastive strategy} to directly model the differences among hierarchical types, and impart the type knowledge from type-rich expressions to predictor.

\end{itemize}

\subsection{Prompt-guided Expression Construction (ProExp)}
\label{sec:ProExp}
ProExp aims to convert the input sentences into type-scarce expressions and type-rich expressions based on entity-oriented prompts \cite{ding2021prompt}. The former could make models sensitive to type patterns in context, and the latter can be taken as the type knowledge resources based on type exposure. 

\paragraph{Type-scarce Expression} 
For each input sentence $x_n$, we construct type-scarce expression $x_n^{ts}$  to guide the pre-trained language model (PLM, e.g. BERT \cite{devlin-etal-2019-bert} used in this paper) encoder to efficiently exploit the entity contextual information, especially the type information. For simplicity, we choose declarative entity-oriented prompts to avoid grammatical errors. 

Specifically, we first copy the marked entity $e_n$ in $x_n$, then add a few conjunctions following the entity. Next, we add two specific words. One of them is \emph{"[MASK]"} at the end of the expression, as a dummy for non-specific type. The other one is \emph{"[ENT]"}, which bridges the original entity expression and prompt, and serves as an entry point for receiving type knowledge from type-rich expressions in ConTKT. The form of $x_n^{ts}$ is as follows:

\begin{quote} \footnotesize 
\centering
\emph{$x_n^{ts}$ = $x_n$} \emph{[ENT] $e_n$ is a [MASK].}
\end{quote}


\begin{figure}[t]\
	\centering
	\includegraphics*[clip=true,width=0.46\textwidth,height=0.24\textheight]{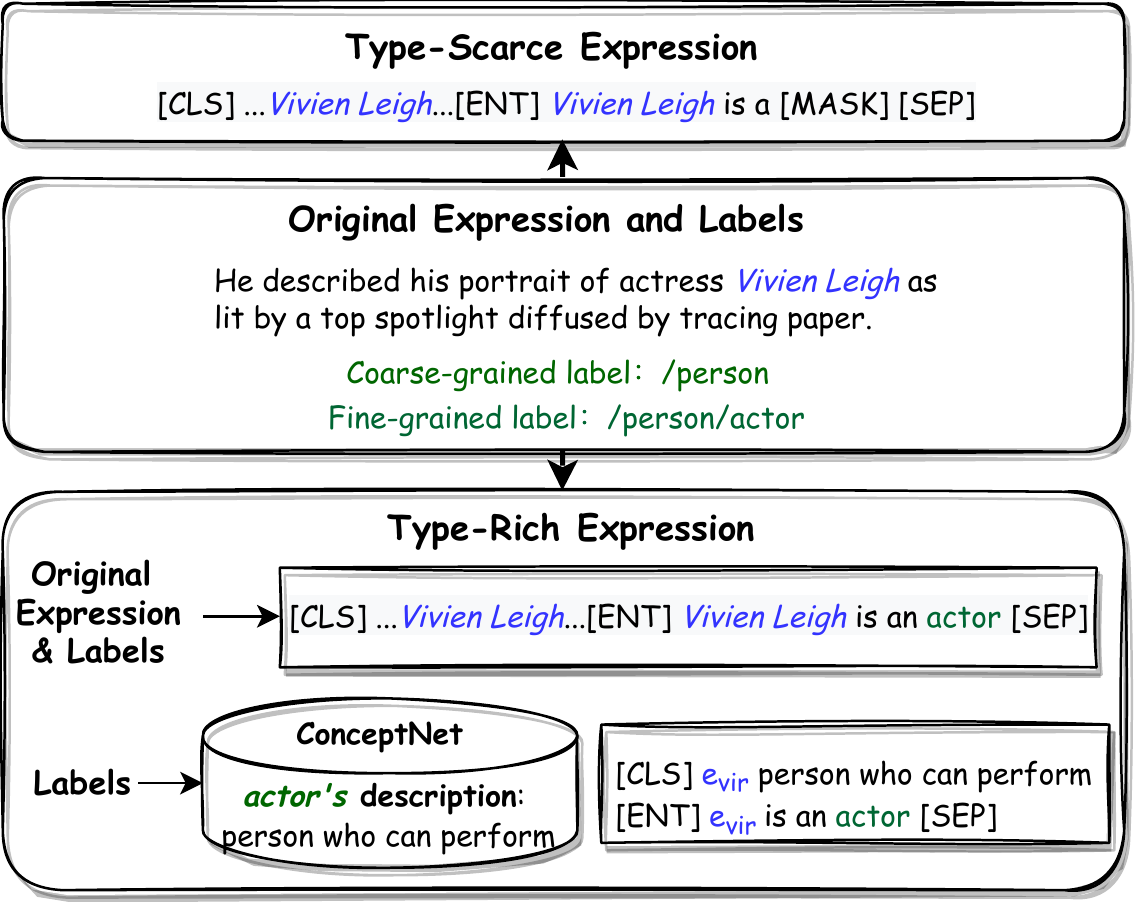}
	\caption{The illustration of prompt-guided expression construction (ProExp, Sec. \ref{sec:ProExp}).} \label{fig3}
\end{figure}

\paragraph{Type-rich Expression}
For each input sentence $x_n$, we also construct two kinds of type-rich expression $x_n^{tr}$  as type knowledge resources for transfer in ConTKT when training. Intuitively, exposing the types directly to the context makes the expressions of entities type-aware. 

Heuristically, fine-grained types contain both coarse- and fine- grained type properties. Therefore, we construct the type-rich expression $x_n^{tr}$ of entity $e_n$ by replacing the dummy type placeholder \emph{"[MASK]"} in its type-scarce expression $x_n^{ts}$ with its fine-grained types in $Y_n$\footnote{For entities with multiple fine-grained types, we concatenate the fine-grained types into one phrase by \emph{"and"}.}. Taking the entity in Figure \ref{fig3} as an example, the form of $x_n^{tr}$ is as follows:

\begin{quote} \footnotesize 
\centering
\emph{$x_n^{tr}$ = $x_n$} \emph{[ENT] $e_n$ is an actor.}
\end{quote}


Moreover, to better show what a particular type is, we introduce several descriptions (2 or 3) of each type from ConceptNet \cite{speer2017conceptnet}. Then we use them to directly expose richer type knowledge to construct extra type-rich expression $x_{type}^{tr}$. Specifically, for each fine-grained type, we replace $x_n$ in $x_n^{tr}$ with the combination of a virtual entity $e_{vir}$ and one of descriptions. Taking the \emph{actor} as an example:

\begin{quote} \footnotesize 
\emph{$x_{actor}^{tr}$ =} \emph{$e_{vir}$ person who can perform [ENT]} \colorbox{white}{$\qquad\quad$} \emph{$e_{vir}$ is an actor.}
\end{quote}


\subsection{Contrastive Type Knowledge Transfer (ConTKT)}
\label{sec:ConTKT}
\begin{figure}[t]\
	\centering
	\includegraphics*[clip=true,width=0.46\textwidth,height=0.20\textheight]{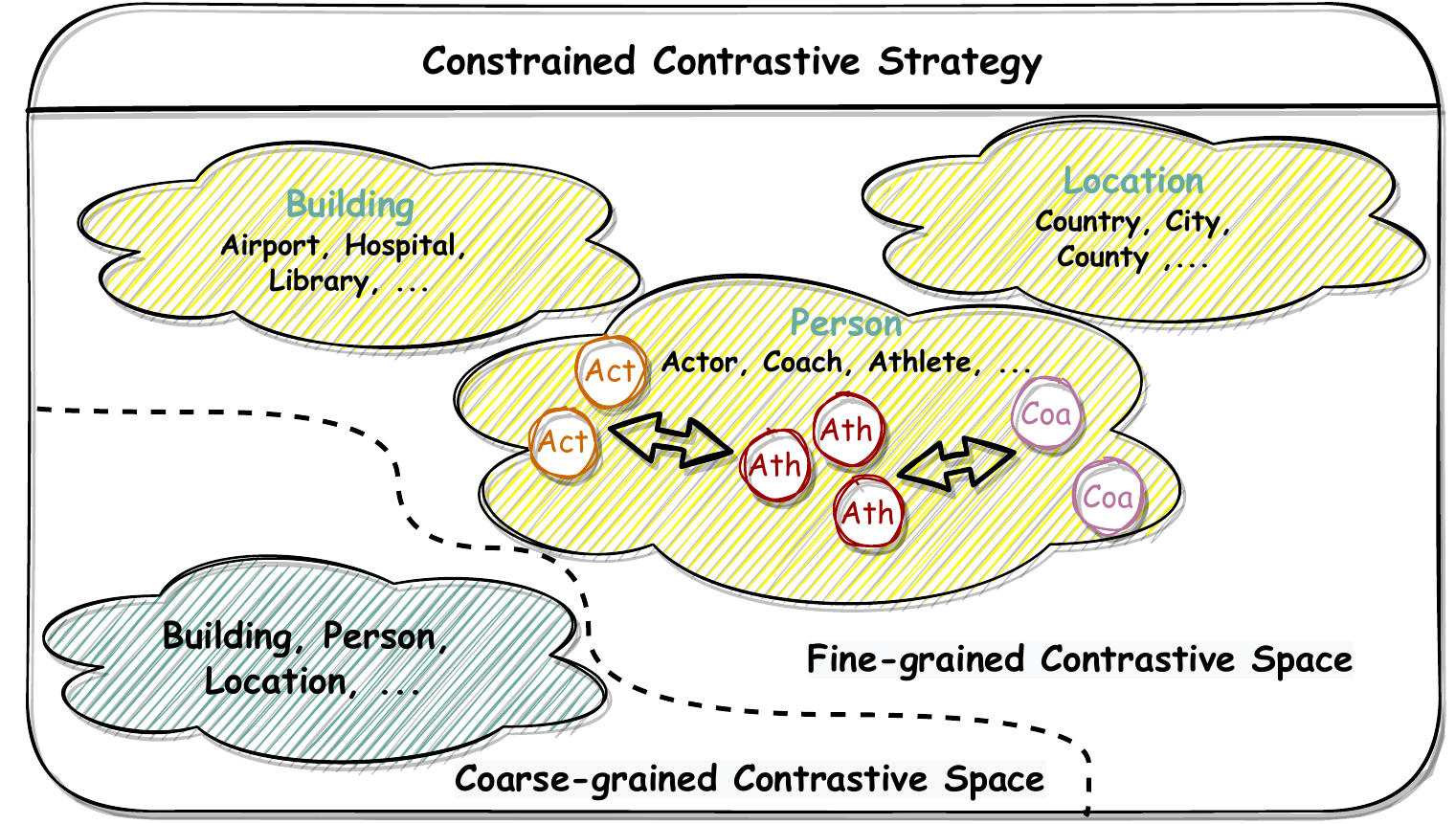}
	\caption{The illustration of constrained contrastive strategy in ConTKT (Sec. \ref{sec:ConTKT}).} \label{fig4}
\end{figure}

ConTKT aims to directly model the differences among hierarchical types, and impart the type knowledge from type-rich expressions to predictor when training.

\paragraph{Expression Encoding}
We design two BERT encoders to encode two kinds of prompt-guided expressions for each entity $e_n$ respectively. One is the encoder $BertEnc_{sca-\theta}$ for encoding type-scarce expression $x_n^{ts}$ when training and prediction, which digs out the type information of $e_n$ in sentence $x_n$. Another is the encoder $BertEnc_{ric-\theta}$, which masters the type knowledge via encoding the type-rich expression $x_n^{tr}$.

Specifically, we first convert the $x_n^{ts}$ and $x_n^{tr}$ to the input sequences of two encoders respectively. Taking the example in Figure \ref{fig3} as following:



\begin{align} \small
\begin{split}
 w_n^{ts} = & [CLS], w^1_n, ..., w^t_n, [ENT], \\ & w^l_n, ..., w^r_n, is, a, [MASK], ., [SEP],
\end{split}
\end{align}
\vspace{-12pt}

\begin{align} \small
\begin{split}
 w_n^{tr} = & [CLS], w^1_n, ..., w^t_n, [ENT], \\ & w^l_n, ..., w^r_n, is, an, actor, ., [SEP], 
\end{split}
\end{align}
\vspace{-12pt}

\begin{align} \small
\begin{split}
 w_{tr}^{actor} = & [CLS], des^1_{actor}, ..., des^t_{actor}, [ENT], \\ & w_{e_{vir}}, is, an, actor, ., [SEP]. 
\end{split}
\end{align}
where the $des^t_{actor}$ is the token of type description and the $w_{e_{vir}}$ is the token of virtual entity $e_{vir}$.

After encoding, the representation $\bm{h}^{[CLS]_{ts}}_n$ of $ w_n^{ts}$ that encodes the contextual information of $x_n^{ts}$ is used by predictor to predict types of entity $e_n$. Additionally, as mentioned above, the representation $\bm{h}^{[ENT]_{tr}}_n$ of $ w_n^{tr}$ is used as the exit of type knowledge contained in $x^{tr}_n$. Accordingly, the representation $\bm{h}^{[ENT]_{ts}}_n$ of $ w_n^{ts}$ is the entrance to receive the type knowledge from $x_n^{tr}$ when training.

\paragraph{Constrained Contrastive Strategy}
We design a constrained contrastive strategy to directly model the hierarchical type differences based on prompt-guided expressions. Based on this, type knowledge is transferred to $BertEnc_{sca-\theta}$ from type-rich expressions by the contrast interaction between types at different granularities and the parameters sharing with $BertEnc_{ric-\theta}$.

Specifically, we only model the fine-grained type differences under the same coarse-grained type by bringing the same types closer while distancing different types. Likewise, for coarse-grained types, we do not care what the fine-grained types are in the same way. In this way, PICOT models the type differences between different granularities by limiting the scope of attention to types.

To specific, for one input batch $\mathcal{B} \subseteq \mathcal{D}$, there are two optimization objectives $\mathcal{L}^{f}_\theta$ and $\mathcal{L}^{c}_\theta$ to model the differences between fine-grained types and coarse-grained types respectively. And each optimization objective consists of two sub-optimization objectives, one for bringing the same types closer ($\mathcal{L}^{f+}_\theta$ and $\mathcal{L}^{c+}_\theta$), while another for distancing different types  ($\mathcal{L}^{f-}_\theta$ and $\mathcal{L}^{c-}_\theta$):

\begin{align} \small
\mathcal{L}^{f+}_\theta & = \frac{1}{|\mathcal{Y}^{f}_\mathcal{B}|} \sum_{y \in \mathcal{Y}^{f}_\mathcal{B}} \frac{1}{2|\mathcal{B}^{f+}|} \sum_{x_i^{y}, x_j^{y} \in \mathcal{B}^{f+}}^{j \neq i} s(x_i^{y}, x_j^{y}), \\
\mathcal{L}^{f-}_\theta & = - \frac{1}{2|\mathcal{B}^{f-}|} \sum_{x_i^{y_i}, x_j^{y_j} \in \mathcal{B}^{f-}}^{j \neq i, y_i \neq y_j} s(x_i^{y_i}, x_j^{y_j}), \\
\mathcal{L}^{c+}_\theta & = \frac{1}{|\mathcal{Y}^{c}_\mathcal{B}|} \sum_{y \in \mathcal{Y}^{c}_\mathcal{B}} \frac{1}{2|\mathcal{B}^{c+}|} \sum_{x_i^{y}, x_j^{y} \in \mathcal{B}^{c+}}^{j \neq i} s(x_i^{y}, x_j^{y}), \\
\mathcal{L}^{c-}_\theta & = - \frac{1}{2|\mathcal{B}^{c-}|} \sum_{x_i^{y_i}, x_j^{y_j} \in \mathcal{B}^{c-}}^{j \neq i, y_i \neq y_j} s(x_i^{y_i}, x_j^{y_j}), 
\end{align}
\begin{align} \small
s(x_i, x_j) = \lg \frac{e^{(dis(\bm{h}^{E}_i, \bm{h}^{E}_j) / \tau)}}{\sum_{x'_i, x'_j \in \mathcal{B*}}^{j \neq i} e^{dis(\bm{h}^{E'}_i, \bm{h}^{E'}_j) / \tau)}},
\end{align}
where, take the $\mathcal{L}^{f}_\theta$ as illustration, $\mathcal{Y}^{f}_\mathcal{B}$ is the set of all fine-grained types in one batch, $y^{f}_\mathcal{B}$ is one of them. And $\mathcal{B}^{f+}$ consists of all $x_n^{ts}$, $x_n^{tr}$ and $x_{type}^{tr}$ with same fine-grained type $y \in \mathcal{Y}^{f}_\mathcal{B}$. Oppositely, $\mathcal{B}^{f-}$ consists of all $x_n^{ts}$, $x_n^{tr}$ and $x_{type}^{tr}$ with same coarse-grained type but different fine-grained types $y_i, y_j \in \mathcal{Y}^{f}_\mathcal{B}$. Moreover, $s$ is the similarity between $x_i$ and $x_j$, $\bm{h}^{E}_i$ and $\bm{h}^{E}_j$ are the $\bm{h}^{[ENT]}_n$ of $x_i$ and $x_j$ after encoding respectively, $dis$ is the $\ell_{2}$-distance function to measure the distance of two representation, $\tau$ is a temperature that adjusts the concentration level and $\mathcal{B*}$ is $\mathcal{B}^{f+}$ or $\mathcal{B}^{f-}$. Likewise, the optimization objectives are similar for $\mathcal{L}^{c}_\theta$.



\paragraph{Learning of FET} 
After transferring, the $\bm{h}^{[CLS]_{ts}}_n$ of $x_n^{ts}$ output by $BertEnc_{sca-\theta}$, which has learned types knowledge, is fed to the predictor to identify the types of the input $e_n$ as following:
\begin{align} \small
    y^*_n \leftarrow MLP(\bm{h}^{[CLS]_{ts}}_n),
\end{align}
where $y^*_n$ is the predicted types of $e_n$ in $x_n$. 

For training of two encoders and predictor, we add the constrained contrastive losses $\mathcal{L}^{f}_\theta$ and $\mathcal{L}^{c}_\theta$ to the classification loss $\mathcal{L}_{\delta}$. Finally, we minimize the $L$ and stochastic gradient update the $\theta$ and $\delta$ as Algorithm \ref{alg1}:
\begin{align} \small
    \mathcal{L}_{\delta} & = BCEWithLogits(y^*_n, y_n), 
\end{align}
\begin{align} \small
    \mathcal{L} = \mathcal{L}_{\delta} & + \lambda_f (\mathcal{L}^{f+}_\theta + \mathcal{L}^{f-}_\theta) \\
    & + \lambda_c (\mathcal{L}^{c+}_{\theta} + \mathcal{L}^{c-}_{\theta}) \nonumber, 
\end{align}
\begin{align} \small
    \theta,\delta & \leftarrow \eta \nabla \mathcal{L},
\end{align}
where, $\lambda_f$ and $\lambda_c$ are the weights of $\mathcal{L}^{f}_\theta$ and $\mathcal{L}^{c}_\theta$ respectively, $\eta$ is the learning rate. 

\begin{algorithm}[t] \footnotesize
	\caption{Learning of PICOT for FET.}
	\begin{algorithmic}[1]
		\Require type-scarce expression $x^{ts}$, type-rich expression $x^{tr}$, extra type-rich expression $x^{tr}_{type}$.
		\Ensure 
		
		\Function {Prompt-Guided Expression Construction}{} 
		\For{each batch $\mathcal{B} \in \mathcal{D}$} 
		\For{input entity $e_n$ with its sentence $x_n$ $\in \mathcal{B}$} 
		\State Construct type-scarce expression $x_n^{ts}$;
		\State Construct type-rich expression $x_n^{tr}$;
		\EndFor
		\\
		\For{each fine-grained $type$ $\in \mathcal{Y}_\mathcal{B}$} 
		\State Construct extra type-rich expression $x_{type}^{tr}$;
		\EndFor
		\EndFor
		\EndFunction  
		\\
		\Function {Contrastive Type Knowledge Transfer}{}
		\For{each batch $\mathcal{B} \in \mathcal{D}$} 
		\For{each fine-grained type $y_f$ in $\mathcal{Y}_\mathcal{B}$} 
		\State Compute $\mathcal{L}^{f}_\theta$ in equation (4) and (5);
		\EndFor
		\\
		\For{each coarse-grained type $y_c$ in $\mathcal{Y}_\mathcal{B}$} 
		\State Compute $\mathcal{L}^{c}_\theta$ in equation (6) and (7);
		\EndFor
		\\
		\State Compute batch classification loss $\mathcal{L}_{\delta}$ in (10);
		\State Compute $\mathcal{L}$ in equation (11);
		\State Stochastic gradient update $\theta$ and $\delta$ in (12);
		\EndFor
		\EndFunction 
	\end{algorithmic}
	\label{alg1}
\end{algorithm}

\section{Experiments}
\label{sec:exp}
\subsection{Experimental Setup}
\label{subsec:exp_set}
\paragraph{Datasets and Evaluation Metrics}
\label{subsubsec:data}
We conduct experiments on three standard FET datasets and follow the version processed and split by \citet{onoe-etal-2021-modeling}. (1) \textbf{BBN} \cite{weischedel2005bbn}, which contains 56 types and each type has a maximum type hierarchy level of 2; (2) \textbf{OntoNotes} \cite{gillick2014context}, which is sampled from the OntoNotes \cite{weischedel2013ontonotes} corpus and re-annotated with 89 types in 3-level hierarchy. Additionally, we ignore the \emph{other} type, which has no obvious meaning, and categorize it into two-level types; (3) \textbf{FIGER} \cite{ling2012fine}, which contains 113 types and each type also has a maximum type hierarchy level of 2. Table \ref{tab1} is the statistics on the coarse-grained and fine-grained types of three datasets. Moreover, we evaluate three datasets using the standard metrics: Macro F1 (Ma-F1) and Micro F1 (Mi-F1).

\paragraph{Parameters Settings}
\label{subsubsec:params}
For a fair comparison, similar to \citet{onoe-etal-2021-modeling}, the BERT encoders are BERT-Large architecture\footnote{\url{https://github.com/google-research/bert}}, which has 24-layers, 1024-hiddens, and 16-heads. For parameters, we set the learning rate of $\eta$ as 8e-6, and set the temperature $\tau$ of the contrastive loss as 0.1 tuned on the development set. Moreover, we also tune the batch size to 96 on the development set. The $\lambda_f$ and $\lambda_c$ are setted as 0.1/0.1/0.1 and 0.1/0.1/0.01/ for three datasets respectively. And we apply the early stop and AdamW gradient strategy to optimize all models. Additionally, to simulate constraints like the previous work \cite{chen-etal-2020-hierarchical,onoe-etal-2021-modeling}, we use the same three simple rules to modify the model's predictions or training data on BBN datasets: (1) dropping \emph{"person"} if \emph{"organization"} exists, (2) dropping \emph{"location"} if \emph{"gpe"} exists, and (3) replacing \emph{"facility"} by \emph{"fac"}, since both the two tags appear in the training set but only \emph{"fac"} in the test set. See Appendix C for more detailed settings.

\begin{table}[t]
\centering
\begin{tabular}{lccc}
\hline
\textbf{Datasets} & \textbf{\#Coarse} & \textbf{\#Fine} & \textbf{\#Fine/Coarse} \\ \hline
BBN               & 17             & 39           & 2.3           \\
OntoNotes         & 20             & 68           & 3.4           \\
FIGER             & 47             & 66           & 1.4           \\ \hline
\end{tabular}
\caption{Statistics on the coarse-grained and fine-grained types of three datasets.}
\label{tab1}

\end{table}
\begin{table*}[t]
\centering
\begin{tabular}{llcclcclcc}
\hline
\hline
\multirow{2}{*}{\textbf{Methods}} &  & \multicolumn{2}{c}{\textbf{BBN}} &  & \multicolumn{2}{c}{\textbf{OntoNotes}} &  & \multicolumn{2}{c}{\textbf{FIGER}} \\ \cline{3-4} \cline{6-7} \cline{9-10} 
                                  &  & \textbf{Ma-F1}      & \textbf{Mi-F1}      &  & \textbf{Ma-F1}         & \textbf{Mi-F1}         &  & \textbf{Ma-F1}       & \textbf{Mi-F1}      \\ \hline
\citet{ren-etal-2016-afet}                                &  & 74.1       & 75.7       &  & 71.1          & 64.7          &  & 69.3        & 66.4        \\
\citet{abhishek-etal-2017-fine}                                &  & 74.1       & 75.7       &  & 68.5          & 63.3          &  & 78.0        & 74.9        \\
\citet{zhang-etal-2018-fine}                                &  & 75.7       & 75.1       &  & 72.1          & 66.5          &  & 78.7        & 75.5        \\
\citet{chen-etal-2020-hierarchical} (exclusive)                               &  & 63.2       & 61.0       &  & 72.4          & 67.2          &  & 82.6        & \textbf{80.8}        \\
\citet{chen-etal-2020-hierarchical} (undefined)                                &  & 79.7       & 80.5       &  & 73.0          & 68.1          &  & 80.5        & 78.1        \\
\citet{lin-ji-2019-attentive}                                &  & 79.3       & 78.1       &  & 82.9*         & 77.3*         &  & 83.0        & 79.8        \\
\citet{onoe-etal-2021-modeling} (vector)                                &  & 78.3       & 78.0       &  & 76.2          & 68.9          &  & 81.6        & 77.0        \\
\citet{onoe-etal-2021-modeling} (box)                               &  & 78.7       & 78.0       &  & 77.3          & 70.9          &  & 79.4        & 75.0        \\
\citet{liu-etal-2021-fine}                                &  & -          & -          &  & 77.6          & 71.8          &  & -           & -           \\ \hline
\textbf{PICOT}                               &  & \textbf{81.8}       & \textbf{82.2}       &  & \textbf{78.7}          & \textbf{72.1}          &  & \textbf{84.7}        & 79.6        \\ \hline
\end{tabular}
\caption{Results on fine-grained entity typing. *: Not directly comparable since large-scale augmentated data is used. The results are tested for significance at the 0.05 level.}
\label{tab2}
\end{table*}

\paragraph{Compared Methods}
\label{subsubsec:comp}
Same as previous methods, we prefer the following models which use the same versions of three datasets and do not rely on large-scale external knowledge or resources as our compared methods\footnote{There are different versions of three datasets exist.}. (1) \textbf{\citet{ren-etal-2016-afet}}, a embedding method which separately models clean and noisy data with type hierarchy; (2) \textbf{\citet{abhishek-etal-2017-fine}}, a neural network model that jointly learns entities and their contexts representation; (3) \textbf{\citet{zhang-etal-2018-fine}}, a neural architecture which leverages both document and sentence level information; (4) \textbf{\citet{chen-etal-2020-hierarchical} (exclusive)}, a classifier for hierarchical FET that embraces ontological structure with exclusive interpretations; (5) \textbf{\citet{chen-etal-2020-hierarchical} (undefined)}, a same classifier as (4) but with different undefined interpretations; (6) \textbf{\citet{lin-ji-2019-attentive}}, a FET model with a novel attention mechanism and a hybrid type classifier; (7) \textbf{\citet{onoe-etal-2021-modeling} (vector)}, a vector-based model for FET; (8) \textbf{\citet{onoe-etal-2021-modeling} (box)}, a box-based model for FET; (9) \textbf{\citet{liu-etal-2021-fine}}, a FET model with extrinsic and intrinsic dependencies between labels. Moreover, all results of compared methods are directly copied from the previous papers.

\begin{table}[t]
\begin{tabular}{lccc}
\hline
\multirow{2}{*}{\textbf{Methods}} & \multicolumn{3}{c}{\textbf{BBN}}                 \\ \cline{2-4} 
                                  & \textbf{Ma-F1} & \textbf{Mi-F1} & $\bm{\nabla}$ \\ \hline
\textbf{PICOT (our)}                     & \textbf{81.8}     & \textbf{82.8}          & -         \\ \hline
w/o Exp.$_{tr_{des}}$                 & 81.6           & 82.2           & -0.2/-0.6      \\
w/o Exp.$_{tr}$                       &    81.4        &    81.7        & -0.4/-1.1       \\
w/o Exp.$_{ts\&tr}$                   &    81.1    	 &       81.5    & -0.7/-1.3    \\ \hline
Previous SOTA         &    79.7        &      80.5     &      -     \\ \hline
\end{tabular}
\caption{Ablation results of the prompt-guided expression (ProExp, Sec. \ref{sec:ProExp}) of FET on BBN. w/o Exp.$_{tr_{des}}$ denotes a varietal PICOT that without extra type-rich expression when training; w/o Exp.$_{tr}$ denotes a varietal PICOT that without all type-rich expressions when training; w/o Exp.$_{ts\&tr}$ denotes a varietal PICOT that without all prompt-guided expressions when training.}
\label{tab4}
\end{table}
\begin{table}[t]
\begin{tabular}{lccc}
\hline
\multirow{2}{*}{\textbf{Methods}} & \multicolumn{3}{c}{\textbf{BBN}}                                     \\ \cline{2-4} 
                                  & \textbf{Ma-F1} & \multicolumn{1}{l}{\textbf{Mi-F1}} & $\bm{\nabla}$ \\ \hline
\textbf{PICOT (our)}                         & \textbf{81.8}           & \textbf{82.8}          & -              \\ \hline
w/o ConTKT$_{coar}$        &     80.0           &       80.4           &     -1.8/-2.4           \\
w/o ConTKT$_{fine}$           & 80.5           & 81.0           & -1.3/-1.8      \\
w/o ConTKT                 &   80.2      &   80.7    &      -1.6/-2.1  \\ \hline
\end{tabular}
\caption{Ablation results of the contrastive type knowledge transfer (ConTKT, Sec. \ref{sec:ConTKT}) of FET on BBN. w/o ConTKT$_{coar}$ denotes a varietal PICOT that removes coarse-grained contrastive loss; w/o ConTKT$_{fine}$ denotes a varietal PICOT that removes fine-grained contrastive loss; w/o ConTKT denotes a varietal PICOT that removes the whole ConTKT.}
\label{tab3}
\end{table}

\subsection{Our Method vs. State-of-the-art Methods}
Table \ref{tab2} shows the results of FET on BBN, OntoNotes, and FIGER. From the results, we can observe that (see Appendix A for more results):

(1) On BBN, our PICOT outperforms all baselines and achieves the best performance on Macro F1 and Micro F1 values, which are 81.8\% and 82.2\%, outperforming the state-of-the-art by a margin of 2.1\% and 1.7\% respectively, which justifies its effectiveness. Moreover, among the three datasets, the BBN dataset has the least training data but the largest boost. This indicates that PICOT can effectively mine type information in limited labeled data by sensing type knowledge and directly modeling the differences between types.

(2) On OntoNotes, compared with the methods without large external data, our PICOT also achieves the best performance on Macro F1 and Micro F1 values, which are 78.7\% and 72.1\%, outperforming by a margin of 1.1\% and 0.3\% respectively. Although OntoNotes has three times more training data than BBN and can provide more type information to compared models, the proposed PICOT can still further improve the performance, which demonstrates the effectiveness of directly modeling type differences.

(3) On FIGER, the largest one in three datasets, our PICOT outperforms the best compared method on Macro-F1 value by a margin of 1.7\%, which further proves the effectiveness of PICOT in mining type differences with labeled data. It is worth noting that FIGER has a slightly lower performance on the Micro-F1 value due to the inconsistent of some test samples, in which only have fine-grained types (e.g., \emph{"/organization/sports\_team"} is present, but \emph{"/organization"} is missing).

\subsection{Effect of Prompt-guided Expression}
We analyze the effect of the prompt-guided expression (ProExp, Sec. \ref{sec:ProExp}) on BBN dataset. As shown in Table \ref{tab4}, from the results, we can observe that: (1) after removing the type-rich expressions (\emph{\textbf{w/o Exp.$_{tr_{des}}$}} and  \emph{\textbf{w/o Exp.$_{tr}$}}), the performance of FET significantly decreases. This proves that exposing the type information directly to the model can bring great help to determine entity types. (2) Comparing the  \emph{\textbf{w/o Exp.$_{tr_{des}}$}} with the  \emph{\textbf{previous SOTA}}, we find that without introducing any external type descriptions, PICOT could also effectively mine type knowledge within the limited labeled data and improve the performance of FET. (3) Comparing the \emph{\textbf{w/o Exp.$_{tr_{des}}$}} with \emph{\textbf{w/o Exp.$_{tr}$}}, just with a small amount of external types descriptions, our PICOT's type knowledge exposure and transfer framework also enhance the performance. (4) \emph{\textbf{w/o Exp.$_{ts\&tr}$}} also achieves good results without any prompt-guided expressions, which also shows the effectiveness of the contrastive transfer strategy.

\begin{figure}[t]\
	\centering
	\includegraphics*[clip=true,width=0.48\textwidth,height=0.24\textheight]{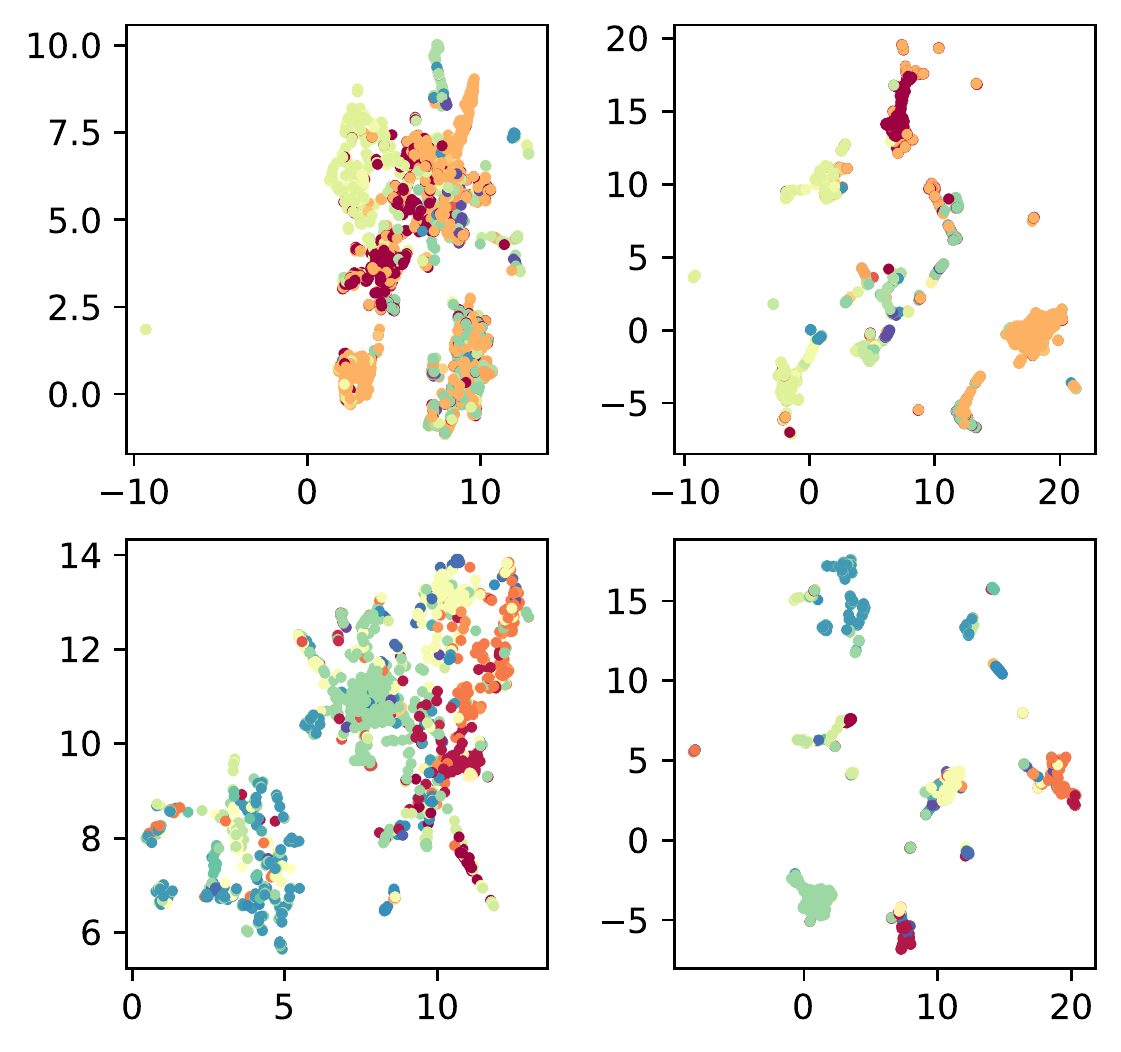}
	\caption{The visualization of type representation clustering without- (left) and with- (right) PICOT on development dataset. Specifically, top row is coarse-grained type clustering and bottom row is fine-grained type clustering. Each color represents a kind of type} \label{fig5}
\end{figure}
\begin{figure}[t]\
	\centering
	\includegraphics*[clip=true,width=0.48\textwidth,height=0.24\textheight]{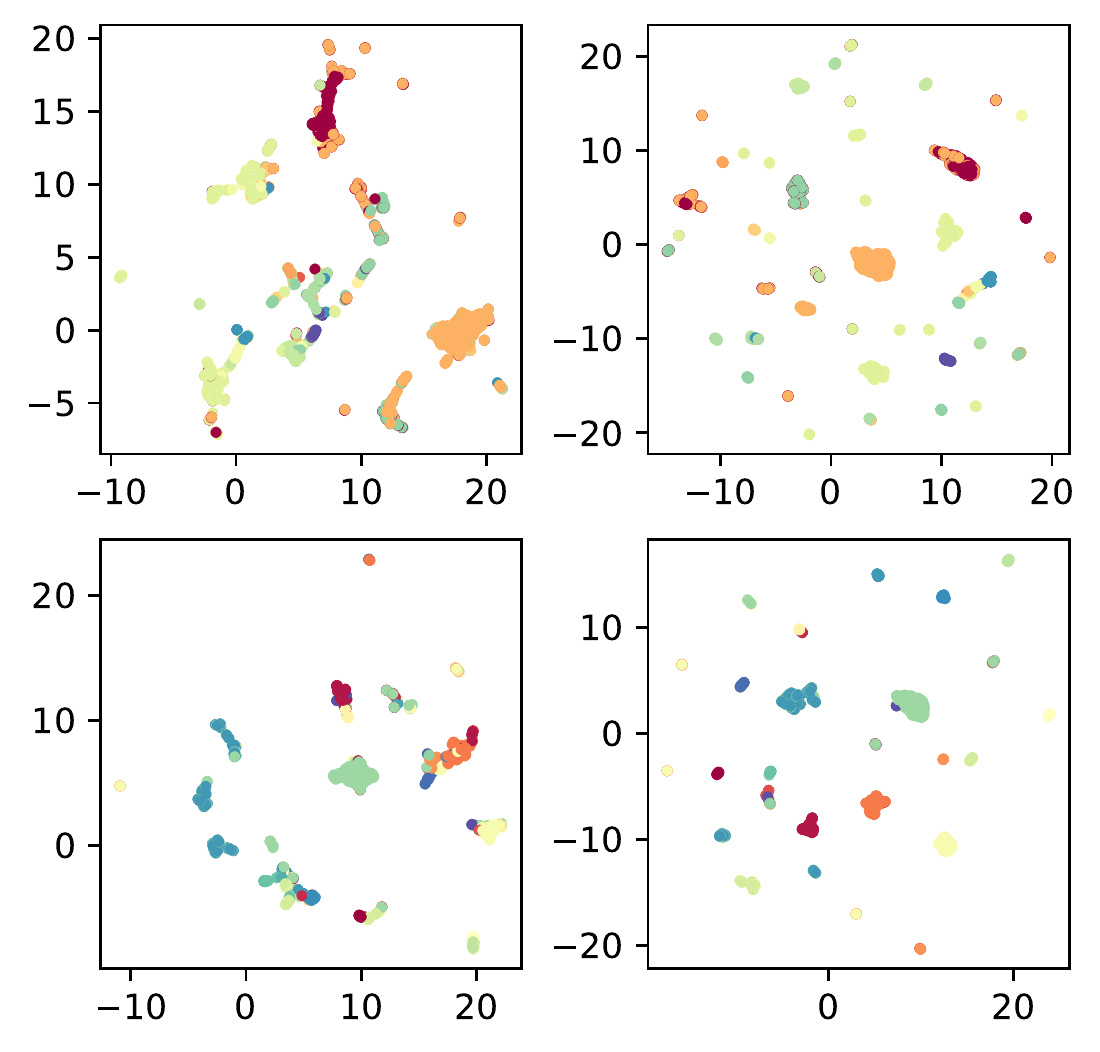}
	\caption{The visualization of type representation clustering of type-scarce (left) and type-rich (right) expressions on development dataset. Specifically, top row is coarse-grained type and bottom row is fine-grained type. Each color represents a kind of type} \label{fig6}
\end{figure}

\subsection{Effect of Contrastive Type Knowledge Transfer}
We analyze the effect of the contrastive type knowledge transfer (ConTKT, Sec. \ref{sec:ConTKT}) on the BBN dataset. As shown in Table \ref{tab3}, from the results, we can observe that: (1) after removing the ConTKT (\emph{\textbf{w/o ConTKT}}), the performance of FET significantly decreases. This illustrates that the contrastive strategy can effectively improve the discrimination of similar types, which is important for FET. (2) Comparing  \emph{\textbf{w/o ConTKT$_{coar}$}},  \emph{\textbf{w/o ConTKT$_{fine}$}} and  \emph{\textbf{PICOT}}, we find that both coarse-grained and fine-grained contrastive training play a key role in the measurement of type differences. (3)  It is worth noting that, comparing  \emph{\textbf{w/o ConTKT$_{coar}$}} with  \emph{\textbf{w/o ConTKT}}, we find that training with type-scarce expression without contrastive strategy works better than only using fine-grained type contrastive strategy. Meanwhile, coarse-grained contrastive training alone (w/o \emph{\textbf{ConTKT$_{fine}$}}) only give a small boost for FET. These indicate that only the combination of coarse-grained and fine-grained contrastive strategies can achieve the desired results. Specifically, coarse-grained contrast ensures the base performance while fine-grained contrast further improves the ability to discriminate types.

\subsection{Visualization of the Effect of Type Distinguishing}

To further illustrate the effect of PICOT, in Figure \ref{fig5}, we cluster the representations of \emph{"[ENT]"} in the type-scarce expressions before and after training by UMAP downscaling \cite{https://doi.org/10.48550/arxiv.1802.03426}. The comparisons of the left and right subgraphs show that the differentiation of \emph{“[ENT]”} representations, which is the entrance for type knowledge, is both greatly improved by PICOT among the coarse-grained and fine-grained types. This illustrates that PICOT can effectively improve the model's ability to discriminate against similar types.

As shown in Figure \ref{fig6}, to elucidate the effect of direct type exposure for type differentiation, we cluster the \emph{"[CLS]"} representations of type-scarce and type-rich expressions, respectively. The comparisons show that the representation of type-rich expressions is more discriminative, especially for fine-grained types, which can effectively guide the model to identify types with high similarity.

\section{Conclusion}
We propose a type-enriched hierarchical contrastive strategy for fine-grained entity typing. Our method can directly model the differences between hierarchical types and improve the ability to distinguish multi-grained similar types. First, we embed types into entity contexts to make type information directly perceptible. Then we design a constrained contrastive strategy on hierarchical taxonomy to directly model type differences at different granularities simultaneously. Experimental results on three benchmarks show that PICOT can achieve state-of-the-art performance on FET with limited annotated data.

\bibliographystyle{acl_natbib}
\bibliography{acl2022}

\begin{thebibliography}{60}
\expandafter\ifx\csname natexlab\endcsname\relax\def\natexlab#1{#1}\fi

\bibitem[{Abhishek et~al.(2017)Abhishek, Anand, and
  Awekar}]{abhishek-etal-2017-fine}
Abhishek Abhishek, Ashish Anand, and Amit Awekar. 2017.
\newblock \href {https://aclanthology.org/E17-1075} {Fine-grained entity type
  classification by jointly learning representations and label embeddings}.
\newblock In \emph{Proceedings of the 15th Conference of the {E}uropean Chapter
  of the Association for Computational Linguistics: Volume 1, Long Papers},
  pages 797--807, Valencia, Spain. Association for Computational Linguistics.

\bibitem[{Chen et~al.(2019)Chen, Gu, Hu, Tang, Hu, Zhuang, and
  Ren}]{chen2019improving}
Bo~Chen, Xiaotao Gu, Yufeng Hu, Siliang Tang, Guoping Hu, Yueting Zhuang, and
  Xiang Ren. 2019.
\newblock Improving distantly-supervised entity typing with compact latent
  space clustering.
\newblock \emph{arXiv preprint arXiv:1904.06475}.

\bibitem[{Chen et~al.(2020{\natexlab{a}})Chen, Kornblith, Norouzi, and
  Hinton}]{pmlr-v119-chen20j}
Ting Chen, Simon Kornblith, Mohammad Norouzi, and Geoffrey Hinton.
  2020{\natexlab{a}}.
\newblock \href {https://proceedings.mlr.press/v119/chen20j.html} {A simple
  framework for contrastive learning of visual representations}.
\newblock In \emph{Proceedings of the 37th International Conference on Machine
  Learning}, volume 119 of \emph{Proceedings of Machine Learning Research},
  pages 1597--1607. PMLR.

\bibitem[{Chen et~al.(2020{\natexlab{b}})Chen, Chen, and
  Van~Durme}]{chen-etal-2020-hierarchical}
Tongfei Chen, Yunmo Chen, and Benjamin Van~Durme. 2020{\natexlab{b}}.
\newblock \href {https://doi.org/10.18653/v1/2020.acl-main.749} {Hierarchical
  entity typing via multi-level learning to rank}.
\newblock In \emph{Proceedings of the 58th Annual Meeting of the Association
  for Computational Linguistics}, pages 8465--8475, Online. Association for
  Computational Linguistics.

\bibitem[{Chen et~al.(2022)Chen, Cheng, Jiang, Liu, Zhang, Shi, and
  Xu}]{chen-etal-2022-learning-sibling}
Yi~Chen, Jiayang Cheng, Haiyun Jiang, Lemao Liu, Haisong Zhang, Shuming Shi,
  and Ruifeng Xu. 2022.
\newblock \href {https://doi.org/10.18653/v1/2022.acl-long.147} {Learning from
  sibling mentions with scalable graph inference in fine-grained entity
  typing}.
\newblock In \emph{Proceedings of the 60th Annual Meeting of the Association
  for Computational Linguistics (Volume 1: Long Papers)}, pages 2076--2087,
  Dublin, Ireland. Association for Computational Linguistics.

\bibitem[{Chen et~al.(2021)Chen, Jiang, Liu, Shi, Fan, Yang, and
  Xu}]{chen-etal-2021-empirical}
Yi~Chen, Haiyun Jiang, Lemao Liu, Shuming Shi, Chuang Fan, Min Yang, and
  Ruifeng Xu. 2021.
\newblock \href {https://doi.org/10.18653/v1/2021.emnlp-main.210} {An empirical
  study on multiple information sources for zero-shot fine-grained entity
  typing}.
\newblock In \emph{Proceedings of the 2021 Conference on Empirical Methods in
  Natural Language Processing}, pages 2668--2678, Online and Punta Cana,
  Dominican Republic. Association for Computational Linguistics.

\bibitem[{Choi et~al.(2018)Choi, Levy, Choi, and
  Zettlemoyer}]{choi-etal-2018-ultra}
Eunsol Choi, Omer Levy, Yejin Choi, and Luke Zettlemoyer. 2018.
\newblock \href {https://doi.org/10.18653/v1/P18-1009} {Ultra-fine entity
  typing}.
\newblock In \emph{Proceedings of the 56th Annual Meeting of the Association
  for Computational Linguistics (Volume 1: Long Papers)}, pages 87--96,
  Melbourne, Australia. Association for Computational Linguistics.

\bibitem[{Dai et~al.(2021)Dai, Song, and Wang}]{dai-etal-2021-ultra}
Hongliang Dai, Yangqiu Song, and Haixun Wang. 2021.
\newblock \href {https://doi.org/10.18653/v1/2021.acl-long.141} {Ultra-fine
  entity typing with weak supervision from a masked language model}.
\newblock In \emph{Proceedings of the 59th Annual Meeting of the Association
  for Computational Linguistics and the 11th International Joint Conference on
  Natural Language Processing (Volume 1: Long Papers)}, pages 1790--1799,
  Online. Association for Computational Linguistics.

\bibitem[{Devlin et~al.(2019)Devlin, Chang, Lee, and
  Toutanova}]{devlin-etal-2019-bert}
Jacob Devlin, Ming-Wei Chang, Kenton Lee, and Kristina Toutanova. 2019.
\newblock \href {https://doi.org/10.18653/v1/N19-1423} {{BERT}: Pre-training of
  deep bidirectional transformers for language understanding}.
\newblock In \emph{Proceedings of the 2019 Conference of the North {A}merican
  Chapter of the Association for Computational Linguistics: Human Language
  Technologies, Volume 1 (Long and Short Papers)}, pages 4171--4186,
  Minneapolis, Minnesota. Association for Computational Linguistics.

\bibitem[{Ding et~al.(2021)Ding, Chen, Han, Xu, Xie, Zheng, Liu, Li, and
  Kim}]{ding2021prompt}
Ning Ding, Yulin Chen, Xu~Han, Guangwei Xu, Pengjun Xie, Hai-Tao Zheng, Zhiyuan
  Liu, Juanzi Li, and Hong-Gee Kim. 2021.
\newblock Prompt-learning for fine-grained entity typing.
\newblock \emph{arXiv preprint arXiv:2108.10604}.

\bibitem[{Fang and Xie(2020)}]{DBLP:journals/corr/abs-2005-12766}
Hongchao Fang and Pengtao Xie. 2020.
\newblock \href {http://arxiv.org/abs/2005.12766} {{CERT:} contrastive
  self-supervised learning for language understanding}.
\newblock \emph{CoRR}, abs/2005.12766.

\bibitem[{Feldman et~al.(2019)Feldman, Davison, and
  Rush}]{DBLP:journals/corr/abs-1909-00505}
Joshua Feldman, Joe Davison, and Alexander~M. Rush. 2019.
\newblock \href {http://arxiv.org/abs/1909.00505} {Commonsense knowledge mining
  from pretrained models}.
\newblock \emph{CoRR}, abs/1909.00505.

\bibitem[{Gillick et~al.(2014)Gillick, Lazic, Ganchev, Kirchner, and
  Huynh}]{gillick2014context}
Dan Gillick, Nevena Lazic, Kuzman Ganchev, Jesse Kirchner, and David Huynh.
  2014.
\newblock Context-dependent fine-grained entity type tagging.
\newblock \emph{arXiv preprint arXiv:1412.1820}.

\bibitem[{He et~al.(2020)He, Fan, Wu, Xie, and Girshick}]{he2020momentum}
Kaiming He, Haoqi Fan, Yuxin Wu, Saining Xie, and Ross Girshick. 2020.
\newblock \href {http://arxiv.org/abs/1911.05722} {Momentum contrast for
  unsupervised visual representation learning}.

\bibitem[{Hjelm et~al.(2018)Hjelm, Fedorov, Lavoie-Marchildon, Grewal, Bachman,
  Trischler, and Bengio}]{hjelm2018learning}
R~Devon Hjelm, Alex Fedorov, Samuel Lavoie-Marchildon, Karan Grewal, Phil
  Bachman, Adam Trischler, and Yoshua Bengio. 2018.
\newblock Learning deep representations by mutual information estimation and
  maximization.
\newblock \emph{arXiv preprint arXiv:1808.06670}.

\bibitem[{Hovy et~al.(2006)Hovy, Marcus, Palmer, Ramshaw, and
  Weischedel}]{hovy2006ontonotes}
Eduard Hovy, Mitch Marcus, Martha Palmer, Lance Ramshaw, and Ralph Weischedel.
  2006.
\newblock Ontonotes: the 90\% solution.
\newblock In \emph{Proceedings of the human language technology conference of
  the NAACL, Companion Volume: Short Papers}, pages 57--60.

\bibitem[{Huang et~al.(2016)Huang, May, Pan, and Ji}]{huang2016building}
Lifu Huang, Jonathan May, Xiaoman Pan, and Heng Ji. 2016.
\newblock Building a fine-grained entity typing system overnight for a new x
  (x= language, domain, genre).
\newblock \emph{arXiv preprint arXiv:1603.03112}.

\bibitem[{Lee et~al.(2006)Lee, Hwang, Oh, Lim, Heo, Lee, Kim, Wang, and
  Jang}]{lee2006fine}
Changki Lee, Yi-Gyu Hwang, Hyo-Jung Oh, Soojong Lim, Jeong Heo, Chung-Hee Lee,
  Hyeon-Jin Kim, Ji-Hyun Wang, and Myung-Gil Jang. 2006.
\newblock Fine-grained named entity recognition using conditional random fields
  for question answering.
\newblock In \emph{Asia information retrieval symposium}, pages 581--587.
  Springer.

\bibitem[{Lester et~al.(2021)Lester, Al{-}Rfou, and
  Constant}]{DBLP:journals/corr/abs-2104-08691}
Brian Lester, Rami Al{-}Rfou, and Noah Constant. 2021.
\newblock \href {http://arxiv.org/abs/2104.08691} {The power of scale for
  parameter-efficient prompt tuning}.
\newblock \emph{CoRR}, abs/2104.08691.

\bibitem[{Leszczynski et~al.(2022)Leszczynski, Fu, Chen, and
  R{\'e}}]{leszczynski2022tabi}
Megan Leszczynski, Daniel~Y Fu, Mayee~F Chen, and Christopher R{\'e}. 2022.
\newblock Tabi: Type-aware bi-encoders for open-domain entity retrieval.
\newblock \emph{arXiv preprint arXiv:2204.08173}.

\bibitem[{Li et~al.(2022)Li, Yin, and Chen}]{li2022ultra}
Bangzheng Li, Wenpeng Yin, and Muhao Chen. 2022.
\newblock Ultra-fine entity typing with indirect supervision from natural
  language inference.
\newblock \emph{arXiv preprint arXiv:2202.06167}.

\bibitem[{Li and Liang(2021)}]{DBLP:journals/corr/abs-2101-00190}
Xiang~Lisa Li and Percy Liang. 2021.
\newblock \href {http://arxiv.org/abs/2101.00190} {Prefix-tuning: Optimizing
  continuous prompts for generation}.
\newblock \emph{CoRR}, abs/2101.00190.

\bibitem[{Lin and Ji(2019)}]{lin-ji-2019-attentive}
Ying Lin and Heng Ji. 2019.
\newblock \href {https://doi.org/10.18653/v1/D19-1641} {An attentive
  fine-grained entity typing model with latent type representation}.
\newblock In \emph{EMNLP-IJCNLP 2019}, pages 6197--6202, Hong Kong, China.
  Association for Computational Linguistics.

\bibitem[{Ling and Weld(2012)}]{ling2012fine}
Xiao Ling and Daniel~S Weld. 2012.
\newblock Fine-grained entity recognition.
\newblock In \emph{Twenty-Sixth AAAI Conference on Artificial Intelligence}.

\bibitem[{Liu et~al.(2021{\natexlab{a}})Liu, Lin, Xiao, Han, Sun, and
  Wu}]{liu-etal-2021-fine}
Qing Liu, Hongyu Lin, Xinyan Xiao, Xianpei Han, Le~Sun, and Hua Wu.
  2021{\natexlab{a}}.
\newblock \href {https://doi.org/10.18653/v1/2021.emnlp-main.378} {Fine-grained
  entity typing via label reasoning}.
\newblock In \emph{Proceedings of the 2021 Conference on Empirical Methods in
  Natural Language Processing}, pages 4611--4622, Online and Punta Cana,
  Dominican Republic. Association for Computational Linguistics.

\bibitem[{Liu et~al.(2021{\natexlab{b}})Liu, Zheng, Du, Ding, Qian, Yang, and
  Tang}]{DBLP:journals/corr/abs-2103-10385}
Xiao Liu, Yanan Zheng, Zhengxiao Du, Ming Ding, Yujie Qian, Zhilin Yang, and
  Jie Tang. 2021{\natexlab{b}}.
\newblock \href {http://arxiv.org/abs/2103.10385} {{GPT} understands, too}.
\newblock \emph{CoRR}, abs/2103.10385.

\bibitem[{Liu and Liu(2021)}]{liu-liu-2021-simcls}
Yixin Liu and Pengfei Liu. 2021.
\newblock \href {https://doi.org/10.18653/v1/2021.acl-short.135} {{S}im{CLS}: A
  simple framework for contrastive learning of abstractive summarization}.
\newblock In \emph{Proceedings of the 59th Annual Meeting of the Association
  for Computational Linguistics and the 11th International Joint Conference on
  Natural Language Processing (Volume 2: Short Papers)}, pages 1065--1072,
  Online. Association for Computational Linguistics.

\bibitem[{L{\'o}pez and Strube(2020)}]{lopez-strube-2020-fully}
Federico L{\'o}pez and Michael Strube. 2020.
\newblock \href {https://doi.org/10.18653/v1/2020.findings-emnlp.42} {A fully
  hyperbolic neural model for hierarchical multi-class classification}.
\newblock In \emph{Findings of the Association for Computational Linguistics:
  EMNLP 2020}, pages 460--475, Online. Association for Computational
  Linguistics.

\bibitem[{Ma et~al.(2016)Ma, Cambria, and Gao}]{ma-etal-2016-label}
Yukun Ma, Erik Cambria, and Sa~Gao. 2016.
\newblock \href {https://aclanthology.org/C16-1017} {Label embedding for
  zero-shot fine-grained named entity typing}.
\newblock In \emph{Proceedings of {COLING} 2016, the 26th International
  Conference on Computational Linguistics: Technical Papers}, pages 171--180,
  Osaka, Japan. The COLING 2016 Organizing Committee.

\bibitem[{McInnes et~al.(2018)McInnes, Healy, and
  Melville}]{https://doi.org/10.48550/arxiv.1802.03426}
Leland McInnes, John Healy, and James Melville. 2018.
\newblock \href {https://doi.org/10.48550/ARXIV.1802.03426} {Umap: Uniform
  manifold approximation and projection for dimension reduction}.

\bibitem[{Murty et~al.(2018)Murty, Verga, Vilnis, Radovanovic, and
  McCallum}]{murty-etal-2018-hierarchical}
Shikhar Murty, Patrick Verga, Luke Vilnis, Irena Radovanovic, and Andrew
  McCallum. 2018.
\newblock \href {https://doi.org/10.18653/v1/P18-1010} {Hierarchical losses and
  new resources for fine-grained entity typing and linking}.
\newblock In \emph{Proceedings of the 56th Annual Meeting of the Association
  for Computational Linguistics (Volume 1: Long Papers)}, pages 97--109,
  Melbourne, Australia. Association for Computational Linguistics.

\bibitem[{Obeidat et~al.(2019)Obeidat, Fern, Shahbazi, and
  Tadepalli}]{obeidat-etal-2019-description}
Rasha Obeidat, Xiaoli Fern, Hamed Shahbazi, and Prasad Tadepalli. 2019.
\newblock \href {https://doi.org/10.18653/v1/N19-1087} {Description-based
  zero-shot fine-grained entity typing}.
\newblock In \emph{Proceedings of the 2019 Conference of the North {A}merican
  Chapter of the Association for Computational Linguistics: Human Language
  Technologies, Volume 1 (Long and Short Papers)}, pages 807--814, Minneapolis,
  Minnesota. Association for Computational Linguistics.

\bibitem[{Onoe et~al.(2021)Onoe, Boratko, McCallum, and
  Durrett}]{onoe-etal-2021-modeling}
Yasumasa Onoe, Michael Boratko, Andrew McCallum, and Greg Durrett. 2021.
\newblock \href {https://doi.org/10.18653/v1/2021.acl-long.160} {Modeling
  fine-grained entity types with box embeddings}.
\newblock In \emph{Proceedings of the 59th Annual Meeting of the Association
  for Computational Linguistics and the 11th International Joint Conference on
  Natural Language Processing (Volume 1: Long Papers)}, pages 2051--2064,
  Online. Association for Computational Linguistics.

\bibitem[{Onoe and Durrett(2020)}]{onoe2020fine}
Yasumasa Onoe and Greg Durrett. 2020.
\newblock Fine-grained entity typing for domain independent entity linking.
\newblock In \emph{Proceedings of the AAAI Conference on Artificial
  Intelligence}, volume~34, pages 8576--8583.

\bibitem[{Pan et~al.(2022)Pan, Wei, and Zhu}]{pan2022automatic}
Weiran Pan, Wei Wei, and Feida Zhu. 2022.
\newblock Automatic noisy label correction for fine-grained entity typing.
\newblock \emph{arXiv preprint arXiv:2205.03011}.

\bibitem[{Pang et~al.(2022)Pang, Zhang, Zhou, and Wang}]{pang-etal-2022-divide}
Kunyuan Pang, Haoyu Zhang, Jie Zhou, and Ting Wang. 2022.
\newblock \href {https://doi.org/10.18653/v1/2022.acl-long.141} {Divide and
  denoise: Learning from noisy labels in fine-grained entity typing with
  cluster-wise loss correction}.
\newblock In \emph{Proceedings of the 60th Annual Meeting of the Association
  for Computational Linguistics (Volume 1: Long Papers)}, pages 1997--2006,
  Dublin, Ireland. Association for Computational Linguistics.

\bibitem[{Petroni et~al.(2019)Petroni, Rockt{\"{a}}schel, Lewis, Bakhtin, Wu,
  Miller, and Riedel}]{DBLP:journals/corr/abs-1909-01066}
Fabio Petroni, Tim Rockt{\"{a}}schel, Patrick S.~H. Lewis, Anton Bakhtin,
  Yuxiang Wu, Alexander~H. Miller, and Sebastian Riedel. 2019.
\newblock \href {http://arxiv.org/abs/1909.01066} {Language models as knowledge
  bases?}
\newblock \emph{CoRR}, abs/1909.01066.

\bibitem[{Rabinovich and Klein(2017)}]{rabinovich-klein-2017-fine}
Maxim Rabinovich and Dan Klein. 2017.
\newblock \href {https://doi.org/10.18653/v1/P17-2052} {Fine-grained entity
  typing with high-multiplicity assignments}.
\newblock In \emph{Proceedings of the 55th Annual Meeting of the Association
  for Computational Linguistics (Volume 2: Short Papers)}, pages 330--334,
  Vancouver, Canada. Association for Computational Linguistics.

\bibitem[{Ren(2020)}]{ren2020fine}
Quan Ren. 2020.
\newblock Fine-grained entity typing with hierarchical inference.
\newblock In \emph{2020 IEEE 4th Information Technology, Networking, Electronic
  and Automation Control Conference (ITNEC)}, volume~1, pages 2552--2558. IEEE.

\bibitem[{Ren et~al.(2016)Ren, He, Qu, Huang, Ji, and Han}]{ren-etal-2016-afet}
Xiang Ren, Wenqi He, Meng Qu, Lifu Huang, Heng Ji, and Jiawei Han. 2016.
\newblock \href {https://doi.org/10.18653/v1/D16-1144} {{AFET}: Automatic
  fine-grained entity typing by hierarchical partial-label embedding}.
\newblock In \emph{Proceedings of the 2016 Conference on Empirical Methods in
  Natural Language Processing}, pages 1369--1378, Austin, Texas. Association
  for Computational Linguistics.

\bibitem[{Schick and Sch{\"u}tze(2021)}]{schick-schutze-2021-exploiting}
Timo Schick and Hinrich Sch{\"u}tze. 2021.
\newblock \href {https://doi.org/10.18653/v1/2021.eacl-main.20} {Exploiting
  cloze-questions for few-shot text classification and natural language
  inference}.
\newblock In \emph{Proceedings of the 16th Conference of the European Chapter
  of the Association for Computational Linguistics: Main Volume}, pages
  255--269, Online. Association for Computational Linguistics.

\bibitem[{Shimaoka et~al.(2017)Shimaoka, Stenetorp, Inui, and
  Riedel}]{shimaoka-etal-2017-neural}
Sonse Shimaoka, Pontus Stenetorp, Kentaro Inui, and Sebastian Riedel. 2017.
\newblock \href {https://aclanthology.org/E17-1119} {Neural architectures for
  fine-grained entity type classification}.
\newblock In \emph{Proceedings of the 15th Conference of the {E}uropean Chapter
  of the Association for Computational Linguistics: Volume 1, Long Papers},
  pages 1271--1280, Valencia, Spain. Association for Computational Linguistics.

\bibitem[{Shin et~al.(2020)Shin, Razeghi, IV, Wallace, and
  Singh}]{DBLP:journals/corr/abs-2010-15980}
Taylor Shin, Yasaman Razeghi, Robert L.~Logan IV, Eric Wallace, and Sameer
  Singh. 2020.
\newblock \href {http://arxiv.org/abs/2010.15980} {Autoprompt: Eliciting
  knowledge from language models with automatically generated prompts}.
\newblock \emph{CoRR}, abs/2010.15980.

\bibitem[{Speer et~al.(2017)Speer, Chin, and Havasi}]{speer2017conceptnet}
Robyn Speer, Joshua Chin, and Catherine Havasi. 2017.
\newblock Conceptnet 5.5: An open multilingual graph of general knowledge.
\newblock In \emph{Thirty-first AAAI conference on artificial intelligence}.

\bibitem[{Suresh and Ong(2021)}]{suresh-ong-2021-negatives}
Varsha Suresh and Desmond Ong. 2021.
\newblock \href {https://doi.org/10.18653/v1/2021.emnlp-main.359} {Not all
  negatives are equal: {L}abel-aware contrastive loss for fine-grained text
  classification}.
\newblock In \emph{Proceedings of the 2021 Conference on Empirical Methods in
  Natural Language Processing}, pages 4381--4394, Online and Punta Cana,
  Dominican Republic. Association for Computational Linguistics.

\bibitem[{Tjong Kim~Sang and
  De~Meulder(2003)}]{tjong-kim-sang-de-meulder-2003-introduction}
Erik~F. Tjong Kim~Sang and Fien De~Meulder. 2003.
\newblock \href {https://aclanthology.org/W03-0419} {Introduction to the
  {C}o{NLL}-2003 shared task: Language-independent named entity recognition}.
\newblock In \emph{Proceedings of the Seventh Conference on Natural Language
  Learning at {HLT}-{NAACL} 2003}, pages 142--147.

\bibitem[{Trinh and Le(2018)}]{DBLP:journals/corr/abs-1806-02847}
Trieu~H. Trinh and Quoc~V. Le. 2018.
\newblock \href {http://arxiv.org/abs/1806.02847} {A simple method for
  commonsense reasoning}.
\newblock \emph{CoRR}, abs/1806.02847.

\bibitem[{Weischedel and Brunstein(2005)}]{weischedel2005bbn}
Ralph Weischedel and Ada Brunstein. 2005.
\newblock Bbn pronoun coreference and entity type corpus.
\newblock \emph{Linguistic Data Consortium, Philadelphia}, 112.

\bibitem[{Weischedel et~al.(2013)Weischedel, Palmer, Marcus, Hovy, Pradhan,
  Ramshaw, Xue, Taylor, Kaufman, Franchini et~al.}]{weischedel2013ontonotes}
Ralph Weischedel, Martha Palmer, Mitchell Marcus, Eduard Hovy, Sameer Pradhan,
  Lance Ramshaw, Nianwen Xue, Ann Taylor, Jeff Kaufman, Michelle Franchini,
  et~al. 2013.
\newblock Ontonotes release 5.0 ldc2013t19.
\newblock \emph{Linguistic Data Consortium, Philadelphia, PA}, 23.

\bibitem[{Wu et~al.(2019)Wu, Zhang, Mao, Guo, and Huai}]{wu2019modeling}
Junshuang Wu, Richong Zhang, Yongyi Mao, Hongyu Guo, and Jinpeng Huai. 2019.
\newblock Modeling noisy hierarchical types in fine-grained entity typing: A
  content-based weighting approach.
\newblock In \emph{IJCAI}, pages 5264--5270.

\bibitem[{Xiong et~al.(2019)Xiong, Wu, Lei, Yu, Chang, Guo, and
  Wang}]{xiong-etal-2019-imposing}
Wenhan Xiong, Jiawei Wu, Deren Lei, Mo~Yu, Shiyu Chang, Xiaoxiao Guo, and
  William~Yang Wang. 2019.
\newblock \href {https://doi.org/10.18653/v1/N19-1084} {Imposing
  label-relational inductive bias for extremely fine-grained entity typing}.
\newblock In \emph{Proceedings of the 2019 Conference of the North {A}merican
  Chapter of the Association for Computational Linguistics: Human Language
  Technologies, Volume 1 (Long and Short Papers)}, pages 773--784, Minneapolis,
  Minnesota. Association for Computational Linguistics.

\bibitem[{Xu and Barbosa(2018)}]{xu2018neural}
Peng Xu and Denilson Barbosa. 2018.
\newblock Neural fine-grained entity type classification with hierarchy-aware
  loss.
\newblock \emph{arXiv preprint arXiv:1803.03378}.

\bibitem[{Xu et~al.(2020)Xu, Guo, Tang, Su, Shou, Gong, Zhong, Quan, Duan, and
  Jiang}]{xu2020syntax}
Zenan Xu, Daya Guo, Duyu Tang, Qinliang Su, Linjun Shou, Ming Gong, Wanjun
  Zhong, Xiaojun Quan, Nan Duan, and Daxin Jiang. 2020.
\newblock Syntax-enhanced pre-trained model.
\newblock \emph{arXiv preprint arXiv:2012.14116}.

\bibitem[{Xu et~al.(2021)Xu, Guo, Tang, Su, Shou, Gong, Zhong, Quan, Jiang, and
  Duan}]{xu-etal-2021-syntax}
Zenan Xu, Daya Guo, Duyu Tang, Qinliang Su, Linjun Shou, Ming Gong, Wanjun
  Zhong, Xiaojun Quan, Daxin Jiang, and Nan Duan. 2021.
\newblock \href {https://doi.org/10.18653/v1/2021.acl-long.420}
  {Syntax-enhanced pre-trained model}.
\newblock In \emph{Proceedings of the 59th Annual Meeting of the Association
  for Computational Linguistics and the 11th International Joint Conference on
  Natural Language Processing (Volume 1: Long Papers)}, pages 5412--5422,
  Online. Association for Computational Linguistics.

\bibitem[{Zhang et~al.(2021{\natexlab{a}})Zhang, Nan, Wei, Li, Zhu, McKeown,
  Nallapati, Arnold, and Xiang}]{DBLP:journals/corr/abs-2103-12953}
Dejiao Zhang, Feng Nan, Xiaokai Wei, Shang{-}Wen Li, Henghui Zhu, Kathleen~R.
  McKeown, Ramesh Nallapati, Andrew~O. Arnold, and Bing Xiang.
  2021{\natexlab{a}}.
\newblock \href {http://arxiv.org/abs/2103.12953} {Supporting clustering with
  contrastive learning}.
\newblock \emph{CoRR}, abs/2103.12953.

\bibitem[{Zhang et~al.(2021{\natexlab{b}})Zhang, Long, Xu, Zhu, Xie, Huang, and
  Wang}]{zhang2021learning}
Haoyu Zhang, Dingkun Long, Guangwei Xu, Muhua Zhu, Pengjun Xie, Fei Huang, and
  Ji~Wang. 2021{\natexlab{b}}.
\newblock Learning with noise: improving distantly-supervised fine-grained
  entity typing via automatic relabeling.
\newblock In \emph{Proceedings of the Twenty-Ninth International Conference on
  International Joint Conferences on Artificial Intelligence}, pages
  3808--3815.

\bibitem[{Zhang et~al.(2018)Zhang, Duh, and Van~Durme}]{zhang-etal-2018-fine}
Sheng Zhang, Kevin Duh, and Benjamin Van~Durme. 2018.
\newblock \href {https://doi.org/10.18653/v1/S18-2022} {Fine-grained entity
  typing through increased discourse context and adaptive classification
  thresholds}.
\newblock In \emph{Proceedings of the Seventh Joint Conference on Lexical and
  Computational Semantics}, pages 173--179, New Orleans, Louisiana. Association
  for Computational Linguistics.

\bibitem[{Zhang et~al.(2020)Zhang, Xia, Lu, and Yu}]{zhang-etal-2020-mzet}
Tao Zhang, Congying Xia, Chun-Ta Lu, and Philip Yu. 2020.
\newblock \href {https://doi.org/10.18653/v1/2020.coling-main.7} {{MZET}:
  Memory augmented zero-shot fine-grained named entity typing}.
\newblock In \emph{Proceedings of the 28th International Conference on
  Computational Linguistics}, pages 77--87, Barcelona, Spain (Online).
  International Committee on Computational Linguistics.

\bibitem[{Zhou et~al.(2018)Zhou, Khashabi, Tsai, and
  Roth}]{zhou-etal-2018-zero}
Ben Zhou, Daniel Khashabi, Chen-Tse Tsai, and Dan Roth. 2018.
\newblock \href {https://doi.org/10.18653/v1/D18-1231} {Zero-shot open entity
  typing as type-compatible grounding}.
\newblock In \emph{Proceedings of the 2018 Conference on Empirical Methods in
  Natural Language Processing}, pages 2065--2076, Brussels, Belgium.
  Association for Computational Linguistics.

\bibitem[{Zuo et~al.(2021)Zuo, Cao, Chen, Liu, Zhao, Peng, and
  Chen}]{zuo-etal-2021-improving}
Xinyu Zuo, Pengfei Cao, Yubo Chen, Kang Liu, Jun Zhao, Weihua Peng, and Yuguang
  Chen. 2021.
\newblock \href {https://doi.org/10.18653/v1/2021.findings-acl.190} {Improving
  event causality identification via self-supervised representation learning on
  external causal statement}.
\newblock In \emph{Findings of the Association for Computational Linguistics:
  ACL-IJCNLP 2021}, pages 2162--2172, Online. Association for Computational
  Linguistics.

\end{thebibliography}

\newpage
\appendix
\section{Supplementary Experiment Results}
\label{sec:SupExp}
\subsection{Effect of the Weights of Different Contrastive Loss}

\begin{table}[h]
\centering
\begin{tabular}{c|ccc}
\hline
\textbf{\begin{tabular}[c]{@{}c@{}}Coarse\\ ---------\\ Fine\end{tabular}} & \textbf{0.01} & \textbf{0.1}                      & \textbf{0.5} \\ \hline
\textbf{0.01}                                                              & -    & \multicolumn{1}{l}{81.6} & -   \\
\textbf{0.1}                                                               & 81.4          & \textbf{81.8}                              & 81.3         \\
\textbf{0.5}                                                               & -             & 78.5                              & -            \\ \hline        
\end{tabular}
\caption{Macro F1 of PICOT on BBN with different coarse- ($\lambda_c$) and fine-grained ($\lambda_f$) contrastive weights.}
\label{tab5}
\end{table}

\begin{table}[h]
\centering
\begin{tabular}{c|ccc}
\hline
\textbf{\begin{tabular}[c]{@{}c@{}}Coarse\\ ---------\\ Fine\end{tabular}} & \textbf{0.01} & \textbf{0.1}             & \textbf{0.5} \\ \hline
\textbf{0.01}                                                              & -             & \multicolumn{1}{l}{82.1} & -            \\
\textbf{0.1}                                                               & 82.0          & \textbf{82.8}            & 81.8         \\
\textbf{0.5}                                                               & -             & 79.9                     & -            \\ \hline
\end{tabular}
\caption{Micro F1 of PICOT on BBN with different coarse- ($\lambda_c$) and fine-grained ($\lambda_f$) contrastive weights.}
\label{tab6}
\end{table}

To further explore the effect of contrastive loss of different granularities, we vary the weights of fine- and coarse-grained contrastive loss to observe the performance of PICOT on the BBN test set, respectively. As shown in Table \ref{tab5} and \ref{tab6}, we can notice that the type knowledge is not fully migrated when the coarse-grained and fine-grained contrastive loss weights are too small, and overly affects the classification performance when the weights are too large. It is worth noting that excessive fine-grained contrastive loss weights significantly degrade the performance because many fine-grained types are not completely distinct, and some types could occur simultaneously. Therefore, excessive differentiation of fine-grained types will confuse models.

\subsection{Effect of the \emph{"[ENT]"} Position}
\begin{table}[h]
\centering
\begin{tabular}{c|cc}
\hline
\textbf{Pos.}            & \textbf{Ma-F1} & \textbf{Mi-F1} \\ \hline
\textbf{After \emph{"{[}CLS{]}}"} & 80.7  & 81.2  \\
\textbf{Before prompt}   & \textbf{81.8}  & \textbf{82.8} \\ \hline
\end{tabular}
\caption{Performance of PICOT with different \emph{"[ENT]"} positions.}
\label{tab7}
\end{table}

As shown in Table \ref{tab7}, we further explore the effect of the position of \emph{"[ENT]"} as type knowledge exit and entry in the input sequence on the PICOT performance. From the results, we can see that placing \emph{"[ENT]"} between the entity context and the type prompt allows for more efficient migration and reception of type knowledge.

\section{Supplementary Related Work}
\label{sec:SupRel}
Named entity recognition \cite{tjong-kim-sang-de-meulder-2003-introduction} and entity typing \cite{ling2012fine,gillick2014context} are fundamental research problems in NLP. Recently researchers pay more attention on fine-grained entity typing (FET) and ultra-fine entity typing (UFET) \cite{choi-etal-2018-ultra}, which predicts specific fine or ultra-fine types for given entities. To do so, obtaining more labeled data is the first research perspective for FET, represented by the distant supervision annotation method \cite{ling2012fine}. With these, some researches had focused on how to reduce noises in automatically labeled data, such as a heuristic constraint pruning approach \cite{gillick2014context}, a partial-label loss \cite{ren-etal-2016-afet}, a penalty optimization term \cite{ren2020fine}, and a novel content-sensitive weighting schema \cite{wu2019modeling}. 

Additionally, one key challenge is how to deal with hierarchical type ontology. Most prior works regarded the hierarchical typing problem as a multi-label classification task and incorporated the hierarchical structure in different ways. \citet{ren-etal-2016-afet} used a predefined label hierarchy to reduce noises; \citet{shimaoka-etal-2017-neural} proposed a hierarchical label encoding method; \citet{xu2018neural} employed a normalized hierarchical loss; \citet{murty-etal-2018-hierarchical} learned a subtyping relation to constrain the type embedding; \citet{chen-etal-2020-hierarchical} designed a novel loss function to exploit label hierarchies. 

Some work attempted to mine more label information or better label representation. \citet{abhishek-etal-2017-fine} enhanced the label representation by sharing parameters; \citet{lopez-strube-2020-fully} embed types into a high-dimension; \citet{xiong-etal-2019-imposing} introduced associated labels to enhance the label representation;  \citet{rabinovich-klein-2017-fine} exploited co-occurrence structures during label set prediction; \cite{lin-ji-2019-attentive} reconstructed the co-occurrence structure via latent label representation; \citet{liu-etal-2021-fine} reasoned fine-grained types by discovering label dependencies knowledge. Additionally, several novel textual representations were applied to obtain richer entity contextual information. \citet{ding2021prompt} investigated the application of prompt-learning to predict fine-grained entity types. \citet{onoe-etal-2021-modeling} studied the box embeddings to capture hierarchies of types. 

Moreover, FET and UFET suffer from an obvious issue of the unseen types due to the lack of annotated data. Therefore, a variety of paradigms were being studied to alleviate this issue, such as a hierarchical clustering model \cite{huang2016building}, a prototypical embedding method \cite{ma-etal-2016-label}, a context-description matching model based on type descriptions from Wikipedia \cite{obeidat-etal-2019-description}, a classifier based on Freebase types of its type-compatible, \cite{zhou-etal-2018-zero}, a novel framework which transfers the knowledge from seen types to the unseen ones \cite{zhang-etal-2020-mzet}, and an empirical study on multiple auxiliary information \cite{chen-etal-2021-empirical}. To further alleviate the lack of annotated data, some work draws on different large-scale external data or knowledge to understand the types of entities in the sentences. \citet{onoe2020fine} used hyperlinked mentions in Wikipedia to distantly label large scale data and train an entity typing model; \citet{xu-etal-2021-syntax} introduced a new pre-training task of predicting the syntactic distance in dependency tree based on large scale texts; \citet{dai-etal-2021-ultra}  automatically generated new ultra-fine entity typing data with labels; \citet{li2022ultra} presented LITE, a new approach that formulates entity typing as an NLI problem based on external data. 

In summary, few prior works focus on directly modeling the differences between types. Therefore, this paper tries to let models know that one type is different from others without large-scale external resources. 

\section{Main Experimental Environments and Other Parameters Settings}
\label{sec:Params}
\subsection{Experimental Environments} 
We deploy all models on a server with Tesla P40  GPU. Specifically, the configuration environment of the server is ubuntu 16.04, and our framework mainly depends on python 3.8.8 and Torch 1.11.

\subsection{Other Parameters Settings} 
All the final hyper-parameters for evaluation are averaged after 3 independent tunings on the development set. Moreover, the three datasets BBN, OntoNotes, and FIGER achieve optimal results at the 20th/10th/5th epochs, which take half a day, one day, and two days, respectively.


This is an appendix.

\end{document}